\documentclass[journal]{IEEEtran}
\usepackage{cite}
\usepackage{graphicx}
\usepackage{amsmath}
\usepackage{array}
\usepackage{amssymb}
\usepackage{booktabs}
\usepackage{bbding}
\usepackage{subfigure}
\usepackage{textcomp}
\usepackage{todonotes}
\usepackage{gensymb}
\usepackage{xcolor}
\usepackage{color}
\usepackage{url}
\usepackage{multirow}
\usepackage{epsfig}
\usepackage{threeparttable}
\usepackage{algorithmicx,algorithm}
\usepackage{algpseudocode}
\usepackage{bbm}
\usepackage{multicol}
\usepackage{makecell}
\usepackage{graphicx}
\usepackage{subfigure} 
\usepackage{colortbl}
\usepackage{booktabs}

\def\etal{{\em et al.}}

\usepackage{diagbox}
\begin{document}
\title{Domain Generalization for Face Anti-spoofing via Content-aware Composite Prompt Engineering}
\author{
        Jiabao Guo, 
        Ajian Liu$^{\dagger}$, 
        Yunfeng Diao$^{\dagger}$, 
        Jin Zhang, 
        Hui Ma, \\
        Bo Zhao, 
        Richang Hong,~\IEEEmembership{Senior Member,~IEEE,}
        Meng Wang~\IEEEmembership{Fellow,~IEEE}
\thanks{
    Jiabao Guo, Yunfeng Diao, Richang Hong and Meng Wang are with the School of Computer Science, Hefei University of Technology (e-mail: garbo\_guo@hfut.edu.cn, diaoyunfeng@hfut.edu.cn, hongrc.hfut@gmail.com, eric.mengwang@gmail.com). Ajian Liu is with the State Key Laboratory of Multimodal Artificial Intelligence Systems (MAIS), Institute of Automation, Chinese Academy of Sciences (CASIA), Beijing 100190, China (e-mail: ajianliu92@gmail.com). Jin Zhang is with the School of Electronic information Engineering, Taiyuan University of Technology (e-mail:2024511162@link.tyut.edu.cn). Hui Ma is with the School of Computer Science and Engineering, Faculty of Innovation Engineering, Macau University of Science and Technology, Macau, China (e-mail: 3220006153@student.must.edu.mo). Bo Zhao are with the School of Cyber Science and Engineering, Wuhan University, Wuhan, China (e-mail: zhaobo@whu.edu.cn).
    }
    \thanks{
    $\dagger$ Corresponding authors
    }
}
\markboth{}
{\MakeLowercase{~\etal}: Domain Generalization for Face Anti-spoofing via Content-aware
Composite Prompt Engineering}
\maketitle

\begin{abstract}
The challenge of Domain Generalization (DG) in Face Anti-Spoofing (FAS) is the significant interference of domain-specific signals on subtle spoofing clues. Recently, some CLIP-based algorithms have been developed to alleviate this interference by adjusting the weights of visual classifiers. However, our analysis of this class-wise prompt engineering suffers from two shortcomings for DG FAS: (1) The categories of facial categories, such as real or spoof, have no semantics for the CLIP model, making it difficult to learn accurate category descriptions. (2) A single form of prompt cannot portray the various types of spoofing. In this work, instead of class-wise prompts, we propose a novel \textit{Content-aware Composite Prompt Engineering} (\textbf{CCPE}) that generates instance-wise composite prompts, including both fixed template and learnable prompts. Specifically, our CCPE constructs content-aware prompts from two branches: (1) Inherent content prompt explicitly benefits from abundant transferred knowledge from the instruction-based Large Language Model (LLM). (2) Learnable content prompts implicitly extract the most informative visual content via Q-Former. Moreover, we design a Cross-Modal Guidance Module (CGM) that dynamically adjusts unimodal features for fusion to achieve better generalized FAS. Finally, our CCPE has been validated for its effectiveness in multiple cross-domain experiments and achieves state-of-the-art (SOTA) results.
\end{abstract}
\IEEEpeerreviewmaketitle

\section{Introduction}
\IEEEPARstart{F}{ace} Anti-spoofing (FAS) plays a crucial role in safeguarding face recognition systems from presentation attacks (e.g., such as print~\cite{zhang2012face}, replay~\cite{chingovska2012effectiveness} and mask~\cite{liu2022contrastive}). Previous FAS methods~\cite{Liu2018Learning,zhang2020face,yu2020fas,cai2020drl,liu2021face} have shown effectiveness in intra-domain scenarios, but substantial performance degradation occurs when adapting to unseen domains due to interference of domain-specific signals.

\begin{figure}[t]
\centering
\includegraphics[width=1\linewidth]{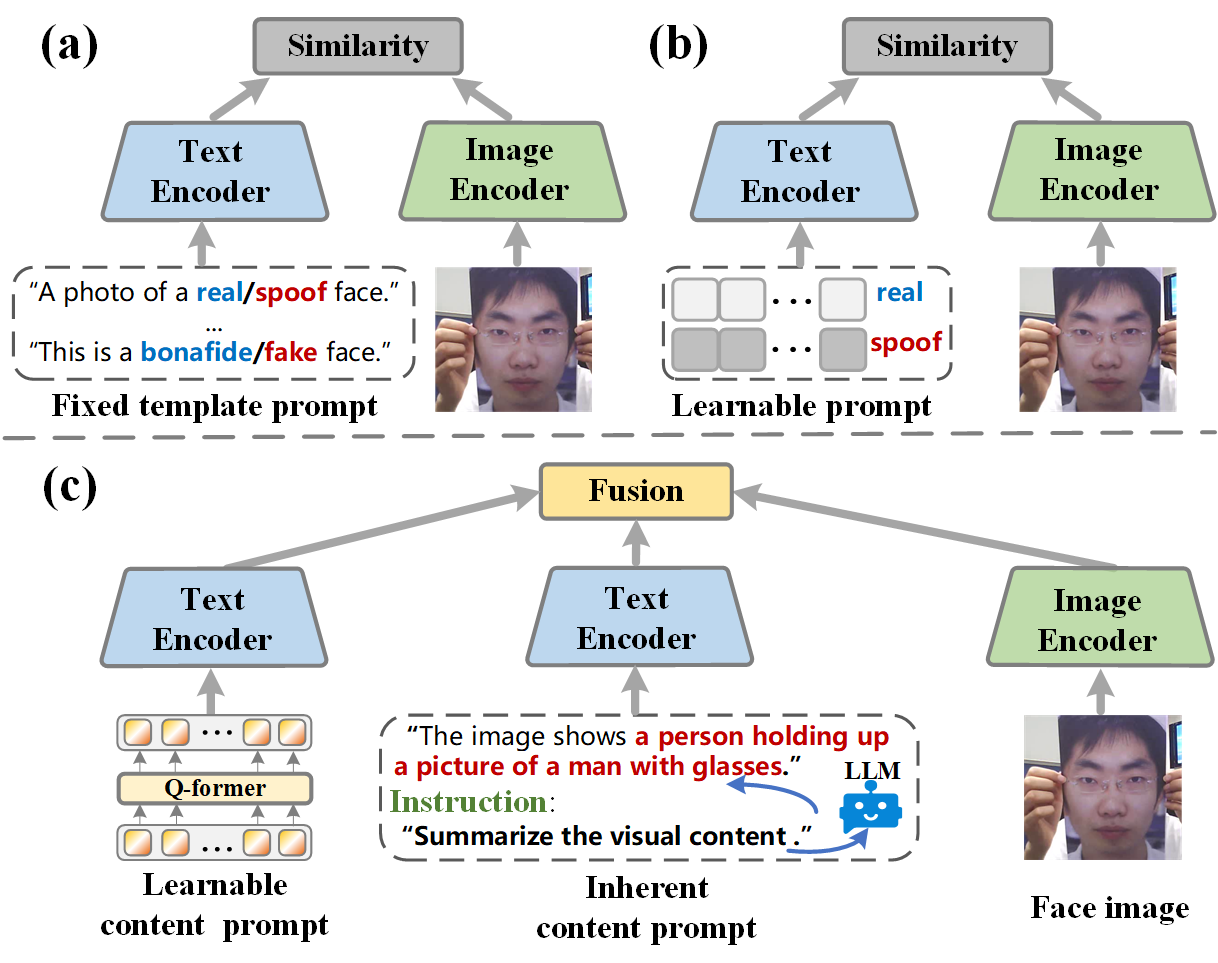}
\caption{Comparison with existing prompt engineering paradigms for DG FAS. (a) Methods based on template prompts, like FLIP~\cite{srivatsan2023flip}, require professional task-specific knowledge to manually design category descriptions. (b) Methods based on prompt learning, like CoOp~\cite{zhou2022coop}, cannot generate accurate category descriptions due to a lack of understanding of their semantics. (c) The proposed CCPE addresses these limitations by constructing both explicit and implicit composite content prompts, which benefit from abundant transferred knowledge by instruction-based LLM and the most informative visual content via Q-former.}
\label{Comparison with existing work}
\end{figure}

To improve generalization, existing methods~\cite{shao2019multi,jia2020single} usually focus on learning domain-invariant representations from facial images. Some approaches aim to acquire a generalized feature space by aligning distributions across multiple domains. However, removing domain-specific signals from the entire face sample features inevitably destroys the semantic structure and leads to restricted generalized features. In recent years, researchers have made use of large-scale vision language models (VLM) such as CLIP~\cite{radford2021learning} by using the textual feature to dynamically adjust the classifier's weights to explore generalizable visual features. Such as FLIP~\cite{srivatsan2023flip} (Fig.~\ref{Comparison with existing work}(a)) establishes multiple natural language class descriptions per class, with the aim of improving FAS generalizability in low-data regimes. To avoid the task-specific knowledge required for manually designing prompt templates, CoOp~\cite{zhou2022coop} (Fig.~\ref{Comparison with existing work}(b)) models a prompt’s context words with learnable vectors while putting the [\textit{CLASS}] (i.e., real and spoof) token in the end position, and CoCoOp~\cite{zhou2022cocoop} further alleviates overfitting in the base classes by learning a lightweight Meta-Net to generate for each image an input-conditional token. However, there are two issues that limit the use of CLIP in DG FAS tasks: (1) The semantic barrier restricts the transfer of knowledge from CLIP. Mechanically, CLIP synthesizes a zero-shot linear classifier for the evaluation of downstream tasks by embedding the names or descriptions of the target dataset classes. While a sample in FAS contains multiple descriptions, such as \textit{`This is a Real/Live face'} for living face, and \textit{`This is a Spoof/Attack/Fake face'} for spoofing face, all of which are meaningless for CLIP~\cite{radford2021learning}. (2) A single form of context prompts cannot depict diverse types of spoofing. Usually, different FAS datasets use different cameras and collection environment settings to collect samples, which makes one form of text unable to describe complex domain-specific signals. Therefore, how to construct semantically valuable and expressive prompts is an important aspect of technology for DG FAS.

For the first issue, based on previous research~\cite{zhang2020face,zhou2023instance}, a complete face sample contains domain-invariant content information and domain-specific signals. Considering that the challenge of DG FAS is the interference of the latter on category attributes, we classify the category of samples as an attribute of content, which not only avoids interference from domain-specific signals but also indirectly induces categories through semantic text description. Therefore, compared to class-wise prompts, instance-wise prompts are more suitable for FAS tasks due to the focus on domain-invariant content features. For the second question, fixed template prompts are good at describing inherent attack types, such as Print, Replay, and 3D Mask, while learnable context prompts can better cover unknown attack types, such as compound attacks. We believe that constructing composite prompts that combine the advantages of both can better solve the difficulty of various attacks in FAS tasks.

In this work, we propose a novel approach called Content-aware Composite Prompt Engineering (CCPE) that generates instance-wise composite prompts, including both fixed template and learnable content prompts. Specifically, instead of class-wise prompt engineering, our CCPE constructs content-aware prompts from two branches: (1) Inherent content prompts explicitly benefit from abundant transferred knowledge from the instruction-based LLM model. (2) Learnable content prompts implicitly extract the most informative visual content via Q-Former. Furthermore, to map the different modalities of vision and language to an aligned representation space, we design a cross-modal guidance module (CGM) that encompasses composite language fusion and vision-language modality fusion. Since multimodal cues tend to be complementary, the fusion of vision-language features as an alternative to textual features as weights for visual features is usually better at capturing different aspects of underlying concepts. In addition, the cross-modal guidance module can dynamically adjust the usage of unimodal features for better generalization of FAS. In this way, generalized visual features are achieved by enhancing the model's understanding of content-aware text semantics. 

In recent years, vision-language pre-trained models have demonstrated outstanding performance in various tasks, where a key point is how to effectively fuse visual and language modalities. Existing methods such as VL-FAS~\cite{fang2024vl} employ text prompts to guide the visual feature extraction process, making it more focused on the facial area, thereby finding more distinct and clear discriminative differences in the samples. The essence of CFPL-FAS~\cite{liu2024cfplfas} is to generate domain-independent prompts through prompt learning, on the one hand, the purpose of learnable prompts is to generate domain-related style prompts, and on the other hand, fixed templates are introduced as supervision signals, endowing task-related semantic information. Unlike constructing prompts before the text encoder, S-CPTL~\cite{guo2024style} uses prompt tokens during the feature extraction process to achieve visual-text semantic alignment. Our CCPE replaces fixed templates with captions generated by large language models, but this approach may lead to ignoring the mutual guidance between modalities to some extent during the feature extraction process. Our proposed method effectively promotes the interaction between text and visual features, making full use of the complementary information from both modalities.

To sum up, the main contributions of this paper are summarized as follows:
\begin{itemize}
\item We propose and formalize a prompt engineering method, called CCPE, laying down the first work to construct composite prompts for DG FAS. It makes use of explicit descriptions of instruction-based LLM and implicit context with learnable prompts.
\item We design a cross-modal guidance module (CGM), which introduces attention mechanism and modality interaction to achieve intra-modal complementarity of information and inter-modal global semantic correlation, respectively.
\item The experimental results show that CCPE consistently achieves greater effectiveness than competing methods.
\end{itemize}

\section{Related Work}
\subsection{Face Anti-Spoofing on Intra-datasets}
In the last decade, FAS methods focusing on intra-dataset scenarios have made significant progress, largely driven by the advent of deep learning techniques, particularly Convolutional Neural Networks (CNNs). Several CNN-based approaches~\cite{Liu2018Learning,liu2022spoof,yu2020fas,cai2020drl,liu2022disentangling} have been proposed to tackle the FAS problem by designing end-to-end frameworks that seamlessly integrate feature extraction and classification stages. These methods harness the powerful representation learning capabilities of CNNs to capture discriminative patterns between live and spoofing faces. Among these works, Liu et al.~\cite{Liu2018Learning} used depth maps and rPPG signals as auxiliary supervision in a CNN-RNN model for face anti-spoofing. Cai et al.~\cite{cai2020drl} enhanced the CNN-RNN framework by introducing deep reinforcement learning to adaptively extract spoofing cues from image sub-patches. They further proposed a learnable network to capture Meta Patterns~\cite{cai2022learning} and integrate them with RGB features for improved performance. Yu et al.~\cite{yu2020fas} introduced a new frame-level FAS method utilizing Central Difference Convolution to capture detailed intrinsic patterns by combining intensity and gradient information from adjacent pixels. However, with the release of high-resolution datasets, that is, OULU-NPU~\cite{Boulkenafet2017OULU}, SiW~\cite{Liu2018Learning}, CelebA-Spoof~\cite{CelebA-Spoof} and high-fidelity mask dataset HiFiMask~\cite{liu2022contrastive}, it is difficult to obtain credible spoofing clues only from the modality of visible light. 

Although algorithms~\cite{george2019biometric,zhang2020casia,liu2021casia} based on multi-modal fusion can alleviate this limitation by leveraging the advantages of information complementarity, they require the testing process to provide modal types consistent with the training phase, which severely limits their deployment in single-modal application scenarios. MA-ViT~\cite{ijcai2022p165} employs early fusion to combine available training modalities, allowing for flexible testing with any given modal samples through a Modality-Agnostic Transformer Block (MATB). FM-ViT~\cite{Liu2023FMViTFM} maintains a dedicated branch for each modality to capture distinct modal information and introduces the Cross-Modal Transformer Block (CMTB), which features multi-headed mutual attention and fusion attention. However, these methods are not specifically designed to address the  problem, which remains a significant challenge in real-world FAS applications. Therefore, researchers are increasingly evaluating the study of flexible modality-based methods~\cite{liu2021face,ijcai2022p165,Liu2023FMViTFM,yu2023visual,yu2023flexiblemodal,liu2024fm}, which aim to improve the performance of a single method utilizing available multi-modal data. 

Despite the promising results achieved by these methods on intra-dataset FAS tasks, their generalization ability to unseen domains remains limited due to variations in imaging conditions, attack types, and sensors across different datasets. As a result, there is increasing interest in creating methods that effectively tackle the domain shift problem in FAS and improve the robustness of FAS systems in real-world applications.

\subsection{Domain Generalizable Face Anti-Spoofing}
As model performance on unseen domains becomes increasingly important, several studies have focused on domain adaptation and  in FAS. Domain Adaptation (DA) methods~\cite{liu2022source,yue2023cyclically,liu2024source,wang2024multi} aim to reduce distribution discrepancies between source and target domains using unlabeled target data. However, obtaining target data during training can be challenging or even impractical. In contrast, Domain Generalization (DG) methods utilize multiple source domains without requiring any target data. Several DG approaches have been proposed for FAS. RFMeta~\cite{shao2020regularized}, SDA~\cite{wang2021self}, AMEL~\cite{zhou2022adaptive}, and D$^{2}$AM~\cite{chen2021generalizable} introduce fine-grained meta-learning frameworks to improve generalization to unseen domains. NAS-FAS~\cite{yu2020fas} employs neural architecture search to identify effective convolution and pooling operators while leveraging cross-domain knowledge. DRDG~\cite{liu2021dual} uses a dual-reweighting strategy to emphasize samples with significant domain bias through sample- and feature-level reweighting. ANRL~\cite{liu2021adaptive} performs adaptive normalized representation learning, imposing dual calibration constraints (inter-domain compatibility and inter-class separability) for better generalization. CIFAS~\cite{liu2022causal} and CRFAS~\cite{zheng2023learning} apply causal intervention to model the data generation process, demonstrating that causal features generalize better across unseen domains. Inspired by theoretical error bounds of DG, NDA~\cite{wang2023domain} and ViT-C\&FA\&CS~\cite{cai2024towards} utilize targeted data augmentation strategies. MADDG\cite{shao2019multi} and SDFANet~\cite{zhou2021selective} employ multiple domain discriminators to balance generalization across diverse image regions and identify common feature representations. MFAE~\cite{zheng2024mfae} proposes a self-supervised pre-training strategy that masks low-frequency information, reducing sensitivity to these features. SSDG-R~\cite{jia2020single} constructs generalized feature spaces by compacting real-face distributions and dispersing fake-face distributions across domains, while maintaining intra-domain compactness. SSAN-R~\cite{wang2022domain} develops a shuffled assembly network to extract and recombine content and style representations within stylized feature spaces. ViTAF~\cite{huang2022adaptive} enhances ViT with adaptive ensemble adapters for FAS, while ViT-DCDCA~\cite{cai2023rehearsal} enables rehearsal-free domain continual learning by addressing both catastrophic forgetting and unseen-domain generalization.
Instead of constructing domain-invariant features, SA-FAS~\cite{sun2023rethinking} enhances domain separability while ensuring consistent transitions from real to spoofed images across domains. IADG~\cite{zhou2023instance} introduces an instance-aware framework to reduce sensitivity to instance-specific patterns, thus improving feature generalization. However, existing methods primarily focus on learning domain-invariant representations from facial images or aligning distributions across multiple domains.

With the rise of vision-language multi-modal, FLIP~\cite{srivatsan2023flip} introduced an end-to-end fine-tuning framework for CLIP, utilizing an ensemble of prompt templates for FAS. CFPL-FAS~\cite{liu2024cfplfas} further advances DG through textual prompt learning. It employs two compact transformer models—Content Q-Former and Style Q-Former—to generate semantic prompts capturing facial image content and style attributes. However, these methods face two primary challenges: a semantic barrier limiting effective knowledge transfer from CLIP, and the insufficient ability of single-form prompts to represent diverse spoofing types. MMGD~\cite{lin2024suppress}, on the other hand, addresses multi-modal DG scenarios by tackling modality unreliability and imbalance through Uncertainty-Guided Cross-Adapter and Rebalanced Modality Gradient Modulation. 

\subsection{Prompt engineering in VLP models} 
Recent advancements in vision-language models (VLMs) highlight the significant potential of textual data in enhancing visual tasks. CLIP~\cite{radford2021learning}, a foundational VLM, leverages large-scale natural language supervision to train image models, enabling effective generalization across various downstream tasks~\cite{peng2023sgva,fang2022transferring}. Besides achieving strong zero-shot performance, CLIP allows task-specific fine-tuning through prompt engineering, further broadening its applicability.

Prompt engineering, originally introduced in natural language processing (NLP), aims to efficiently transfer knowledge from pre-trained language models to downstream tasks. Recent work has extended prompt engineering techniques to VLMs, predominantly by optimizing textual context within CLIP’s language branch. This approach has gained popularity as an efficient alternative to computationally intensive or proprietary text encoders. CoOp~\cite{zhou2022coop} demonstrates that manually crafted prompts significantly influence CLIP's predictions, and proposes replacing fixed templates with learnable contexts adapted through few-shot learning, thereby improving downstream performance. Building upon CoOp, CoCoOp~\cite{zhou2022cocoop} introduces a lightweight Meta-Net to dynamically integrate visual encoder outputs with trainable prompts, enhancing domain adaptability and generalization. KgCoOp~\cite{yao2023visual} further promotes generalization by incorporating general knowledge constraints into prompt embeddings. While these approaches primarily focus on optimizing the language branch of CLIP, MaPLe~\cite{khattak2023maple} proposes multi-modal prompt learning. It introduces a vision-language coupling function that dynamically adapts both modalities' representations, fostering deeper synergy between them. However, using standard classification loss with category labels alone does not always yield sufficiently informative context, motivating our exploration of integrating powerful large language models to leverage richer knowledge.

\begin{figure*}[t]
\centering
\includegraphics[width=0.95\linewidth]{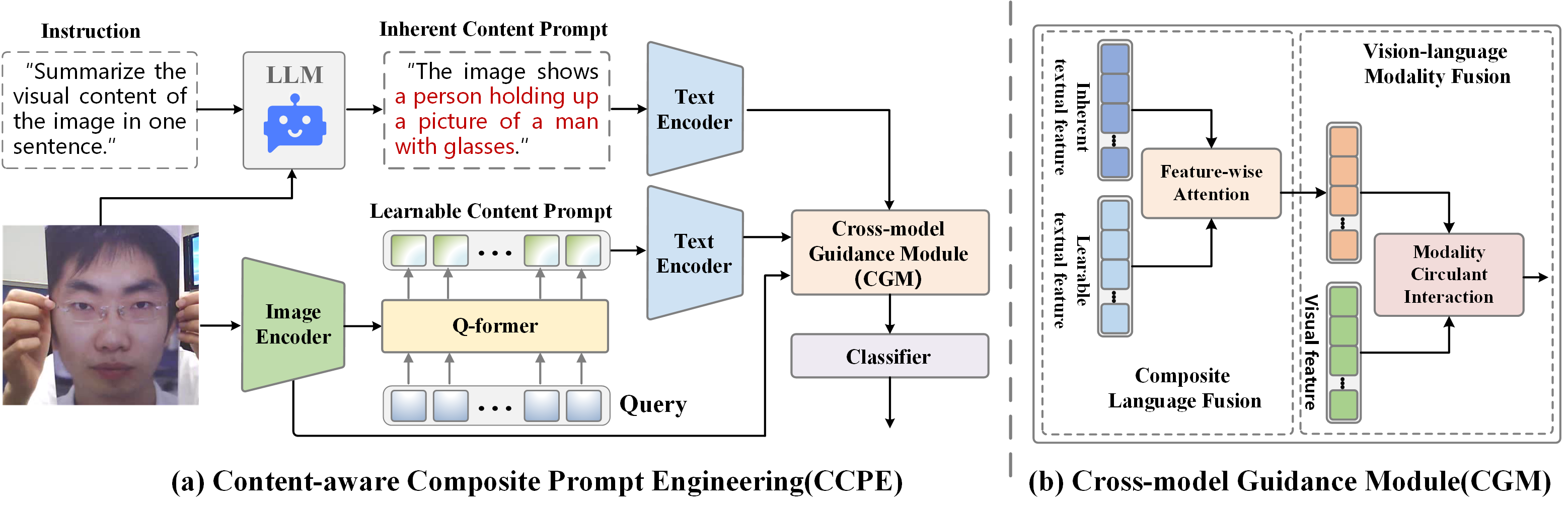}
\caption{\textbf{Overall architecture of our proposed Content-aware Composite Prompt Engineering (CCPE) framework for DG FAS.} Our CCPE is built on CLIP and realizes adaption to FAS tasks by leveraging prompt engineering with two main contributions: \textbf{(1) Content-aware Composite Prompt Engineering.} CCPE generates instance-wise composite prompts, including both inherent content prompts and learnable content prompts. Inherent content prompt explicitly benefits from abundant transferred knowledge from the instruction-based LLM model. Learnable content prompts implicitly extract the most informative visual content via Q-Former. \textbf{(2) Cross-modal Guidance Module (CGM)}. CGM encompasses composite language fusion and vision-language modality fusion. This multimodal-based fusion approach is usually better at capturing different aspects of underlying concepts and dynamically adjusting the usage of unimodal features for better generalization FAS.}
\label{CCPE overview}
\end{figure*}

\section{CCPE: Content-aware Composite Prompt Engineering}
\subsection{Preliminary}

Contrastive Language-Image Pre-training (CLIP) ~\cite{radford2021learning} have demonstrated remarkable capabilities in visual representation learning. This model is uniquely characterized by its dual-encoder structure: an image encoder that converts images into visual embeddings, and a text encoder that transforms textual input into corresponding textual representations.downstream tasks. Formally, we define $\mathbb{D}=\left\{\left(x_i, t_i\right)\right\}_{i=1}^{B}$ as the collection of image-text sample pairs with $C$ categories, and $B$ is the number of images in a mini-batch. $x_i \in \mathbb{R}^{H \times W \times 3}$ is the image and $t_i$ indicates the corresponding text description. The image encoder initially divides the image into $N_v$ fixed-size patches which are projected to create patch embeddings $\boldsymbol{E}_i^p \in \mathbb{R}^{N_v \times d}$, where $d$ is the hidden dimension of CLIP. Each patch embedding, along with a learnable token [CLS], passes sequentially through the transformer blocks to obtain image representation ${v_i}\in \mathbb{R}^{ d}$.

{\flushleft \textbf{Prompt Design.}} In CLIP, prompt tuning adapts the model to downstream tasks by using manually crafted templates to form the prompt while keeping the pre-trained image and text encoders fixed. Given the dataset with category names $\left\{\left[\text{\textit{CLASS}}\right]_c\right\}_{c=1}^C$, each class-wise text description is denoted as $t^{clip}_c = \left\{\text{\textit{A  photo of a} } \left[\text{\textit{CLASS}}\right]_c\right\} $. The text encoder first tokenizes each word of the description $t^{clip}_c$ by assigning a specific numeric ID. Each sequence of tokens, which is enclosed within [SOS] and [EOS] tokens and limited a fixed length of 77, is further project to the word embedding $\boldsymbol{E}_c^w \in \mathbb{R}^{77 \times d}$ and then passed on to the transformer blocks to generate the textual representation $l^ {clip}_c\in \mathbb{R}^d$. Specifically, based on the visual feature $\boldsymbol{v}_{\boldsymbol{i}}$ and textual feature $l^{clip}_c$, the prediction probability is calculated as:

\begin{equation}
p(y=c \mid v_i)=\frac{e^{\operatorname{sim}\left(v_i,\: l^ {clip}_y\right) / \tau}}{\sum_{c=1}^{C} e^{s i m\left(v_i,\: l^ {clip}_c\right) / \tau}},
\end{equation} where $\operatorname{sim}(\cdot)$ refers specifically to cosine similarity and $\tau$ is a temperature parameter. Our method involves the direct utilization of a pre-trained CLIP model. 

\subsection{Overview}

Fig.~\ref{CCPE overview} presents an overall architecture of our proposed Content-aware Composite Prompt Engineering (CCPE) framework for DG FAS. Unlike prior works that endeavor to design category-based prompts, CCPE proposes a content-aware prompting approach where the problem of category names (i.e. real/spoof) without semantics can be solved while the diverse characteristics of categories are guaranteed. Concretely, for a mini-batch of image samples, we first propose a composite prompt engineering architecture that consists of two branches. On the one hand, we use instruction-based LLM to generate inherent content prompt $\boldsymbol{P}_{\mathcal{I}}$. And, on the other hand, the learnable content prompt $ \boldsymbol{P}_{\mathcal{L}}$ can be acquired by extracting the distinct visual information of the instance using the Q-former~\cite{li2023blip}. Then the two branches of the content prompts are set to fixed text encoders and acquire text features $\boldsymbol{t}_{\mathcal{I}}$, $\boldsymbol{t}_{\mathcal{L}}$, respectively. Meanwhile, the image encoder encodes the image samples in visual features $\boldsymbol{V}$. Moreover, in order to learn cross-modal contextual representations, The textual features $\boldsymbol{t}_{\mathcal{I}}$, $\boldsymbol{t}_{\mathcal{L}}$ and visual features  $\boldsymbol{V}$ are fed into a cross-modal guidance module (CGM) after the encoders which include composite language fusion and vision-language modality fusion aiming to achieve intra-modal complementarity of information and inter-modal global semantic correlation, respectively. In the following, we first outline the process of content-aware composite prompt generation in Section \ref{3.3} and then present our proposed cross-modal guidance module in Section \ref{3.4}.

\subsection{Content-aware Composite Prompt Generation}
\label{3.3}

{\flushleft \textbf{Inherent Content Prompt.}} We generate inherent content prompts by adopting "\textit{Summarize the visual content of the image in one sentence.}" to instruct LLM to describe the image content. The description $ \mathbf{T}_i $ would provide specific diverse context information for each face instance. For instance, in the case of generating descriptive words for print attacks in Fig.~\ref{CCPE overview} (a), the instruction-based LLM (i.e. LLaVA~\cite{liu2024visual}) generates the words "\textit{The image shows a person holding up a picture of a man with glasses.}". In general, we devised a conversation between the instruction-based LLM assistant and an individual seeking clarification on the content of this photo. Natural language is used to describe the content of images, enabling it to play a pivotal role in bridging visual signals with linguistic semantics. This process aligns with a prevalent mode of human communication, improving understanding of the intersection between visual perception and language.

Then we use the tokenizer of the pre-trained text encoder to map $\mathbf{T}_i$ into the vocabulary space. This process is formally denoted as $\text{Tokenizer}(\cdot)$. Specifically, we project the squeezes of tokens to the vocabulary space by the pre-trained embedding weights $W_E \in \mathbb{R}^{ 77 \times d} $, which can be formulated as follows:

\begin{equation}
\boldsymbol{P}_{\mathcal{I}} = \left\{\text{Tokenizer}\left( \mathbf{T}_i\right) W_E \right\}_{i=1}^B,  \quad \boldsymbol{P}_{\mathcal{I}} \in \mathbb{R}^{B \times 77 \times d}
\end{equation}
LLaVA, as an advanced open-source tool, merges language generation and image understanding, marking a new frontier in technology. Moreover, the generation of $ \mathbf{T}_i
$ occurs offline before the training phase, thereby preventing any increase in computational demands during training. These advantages make inherent content prompt tailored to enhance FAS in real-world scenarios. 

{\flushleft \textbf{Learnable Content Prompt.}} We employ a Q-Former to learn the most relevant context representations for face images. Formally, given a mini-batch of images $\left\{x_i\right\}_{i=1}^B$, the image encoder $\mathcal{V}(\cdot)$  will first map each image to the visual representation: $\boldsymbol{V}=
\left\{\boldsymbol{v}_i \right\}_{i=1}^B =    \mathcal{V}\left(\left\{x_i\right\}_{i=1}^B\right), \boldsymbol{v}_i \in \mathbb{R}^d$. Then, we randomly initialize $M$ learnable queries $\boldsymbol{Q}=\left\{\boldsymbol{q}_m\right\}_{m=1}^M, \boldsymbol{q}_m \in \mathbb{R}^{d}$. We feed the visual representation, and the queries pass through the Q-former $\mathcal{F}_q(\cdot)$, which shares the same architecture as BLIP-2, to obtain the prompt $ \boldsymbol{P}_{\mathcal{L}}=\left\{\mathbf{q}^{\prime}_m\right\}_{m=1}^M,  \boldsymbol{q}^{\prime}_m \in \mathbb{R}^{d}
$. This process can be expressed as follows:

\begin{equation}
\boldsymbol{P}_{\mathcal{L}} = \left\{\mathcal{F}_q(\boldsymbol{Q}, \boldsymbol{v}_i)\right\}_{i=1}^B, \quad \boldsymbol{P}_{\mathcal{L}} \in \mathbb{R}^{B\times M \times d}
\end{equation}

{\flushleft \textbf{Text Supervision.}} After prompts are tokenized and projected to the word embedding, the inherent content prompt $\boldsymbol{P}_{\mathcal{I}}$ and learnable content prompt $ \boldsymbol{P}_{\mathcal{L}}$  are fed into text encoder to produce enriched features $\left\{ \boldsymbol{t}_{\mathcal{I}}, \boldsymbol{t}_{\mathcal{L}} \right\}$ respectively, which can be formulated as follows:
\begin{equation}
\left\{\boldsymbol{t}_{\mathcal{I}}, \boldsymbol{t}_{\mathcal{L}} \right\} =\left\{\mathcal{T}\left(\boldsymbol{P}_{\mathcal{I}}\right), \mathcal{T}\left(\boldsymbol{P}_{\mathcal{L}}\right)\right\}, \quad \boldsymbol{t}_{\mathcal{I}},\boldsymbol{t}_{\mathcal{L}} \in\mathbb{R}^{B\times d}
\end{equation}
Inherent content prompt prompts are good at describing recognized attack types, while learnable content prompts can better cover unknown attack types. To enforce that specific knowledge features in the fixed description space are close to unified knowledge features, we introduce knowledge-guided Loss: \begin{equation}
\mathcal{L}_{kg}=1- \frac{1}{B} \sum_{i=1}^B \left(\ \cos \left(\boldsymbol{t}_{\mathcal{I}}W_{kg}, \boldsymbol{t}_{\mathcal{L}}  \right)\right)
\end{equation}
To distill knowledge from LLM, we set a monitor factor $ W_{kg} \in \mathbb{R}^{d \times d}$. The cosine distance (denoted as $\cos(\cdot)$) is applied as a regularization to guide the learnable content prompts to be close to their corresponding content-aware representation in the embedding space. The alignment between inherent and learnable content prompts can combine the advantages of both and better solve the difficulty of diverse attacks in FAS tasks. To clarify, the motivation for introducing $\mathcal{L}_{kg}$ loss is to ensure that the learned unknown attacks remain within the attack categories and not to learn text descriptions unrelated to the attack without constraints. Essentially, the loss $\mathcal{L}_{kg}$ is the narrowing of the spatial distance between conventional attack and unknown attack descriptions and does not alter their attack properties. This balance allows the framework to effectively generalize to unseen attack scenarios without sacrificing the richness and variability introduced by composite prompts. So the diversity can still be guaranteed.

\subsection{Cross-modal Guidance Module}
\label{3.4}

\begin{figure}[t]
\centering
\includegraphics[scale=0.45]{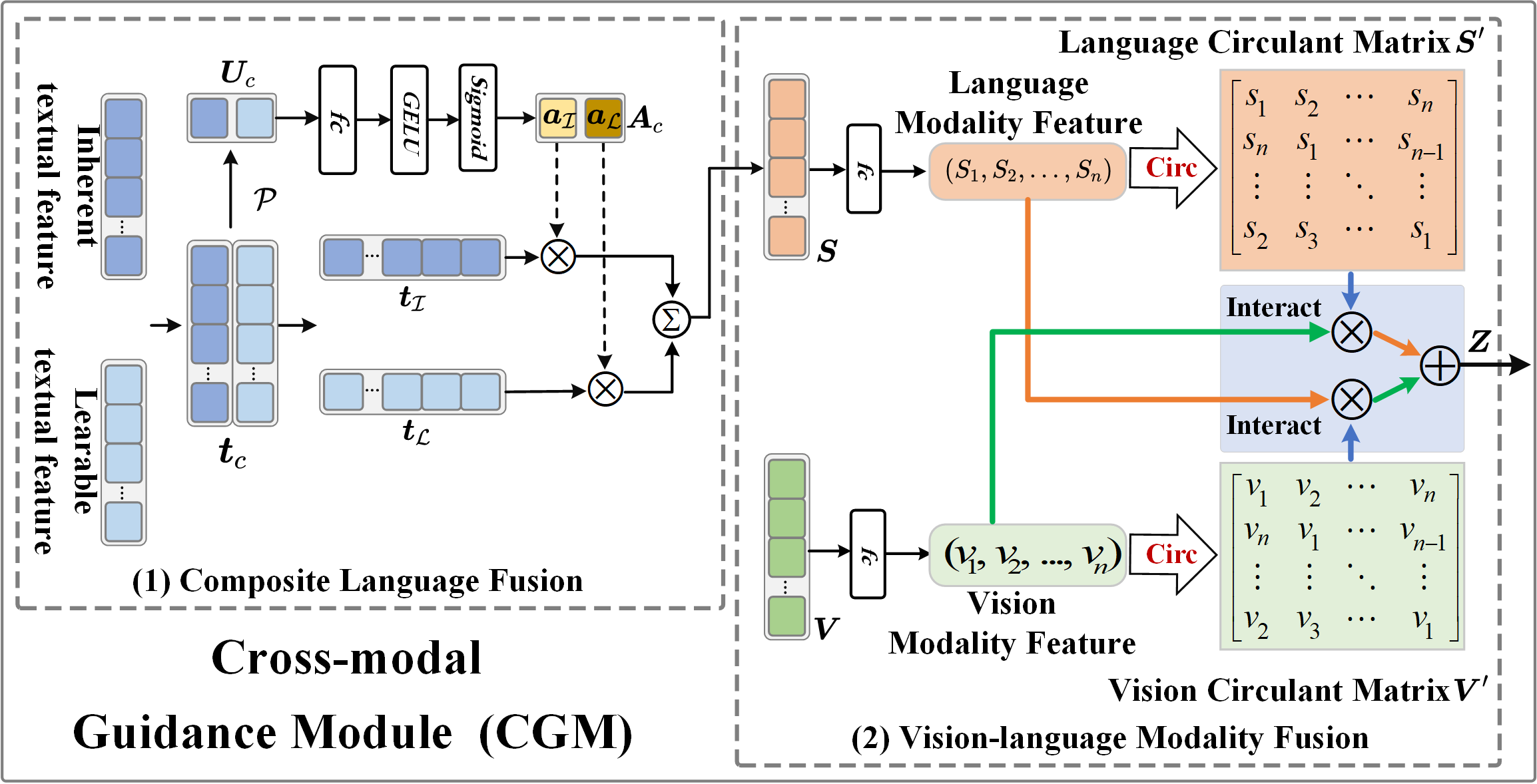}
\caption{The proposed \textbf{Cross-modal Guidance Module (CGM)}. CGM includes two components: composite language fusion and vision-language modality fusion.}
\label{CGM}
\end{figure}

We designed a Cross-modal Guidance Module (CGM)  that aims to mine more potential information under multi-modal content through vision-language feature fusion, rather than assigning textual features as weights for visual features. As depicted in Fig.~\ref{CGM}, CGM achieves composite language fusion and vision-language modality fusion.

{\flushleft \textbf{Composite Language Fusion.}} Aiming to achieve intra-modal complementarity of textual information, we design an attention mechanism to reweight the channels of inherent content prompt representation $\boldsymbol{t}_{\mathcal{I}} \in \mathbb{R}^{B\times d}$ and learnable content prompt representation $\boldsymbol{t}_{\mathcal{L}} \in \mathbb{R}^{B\times d}$ before textual feature aggregation, as shown in Fig.~\ref{CGM}(1). Concretely, we first concatenate the two features to obtain the joint textual feature $\boldsymbol{t}_c\in\mathbb{R}^{B\times d\times 2}$. By applying feature-wise average pooling $\mathcal{P}(.)$, a squeezed vector $\boldsymbol{U}_c \in \mathbb{R}^{B\times 1\times 2}$ is obtained. Then, the initial weight obtained in the previous step is sent into two fully connected layers, denoted $\textit{fc}(\cdot)$. With a \textit{GELU} (denoted as $\sigma(\cdot)$) and \textit{Sigmoid} (denoted as $\delta(\cdot)$) based normalization, we obtain attention weights $ \boldsymbol{A}_c=[\boldsymbol{a}_{\mathcal{I}}\|\boldsymbol{a}_{\mathcal{L}}]_2$ for inherent content prompt and learnable content prompt. The weights are multiplied with corresponding $\boldsymbol{t}_{\mathcal{I}}$ and $\boldsymbol{t}_{\mathcal{L}}$, and a sum process is performed, obtaining a composite language fused feature $\boldsymbol{S} \in \mathbb{R}^{B\times d}$ :
\begin{equation}
\begin{aligned} 
\boldsymbol{t}_c &=[\boldsymbol{t}_{\mathcal{I}}\| \boldsymbol{t}_{\mathcal{L}}]_2, \quad \boldsymbol{t}_c \in \mathbb{R}^{B \times d \times 2 }\\
\boldsymbol{U}_c &= \mathcal{P}(\boldsymbol{t}_c),\quad \boldsymbol{U}_c\in \mathbb{R}^{B\times 1\times 2}\\
\boldsymbol{A}_c &=\delta (\sigma(\textit{fc}(\boldsymbol{U}_c))),\quad \boldsymbol{A}_c\in \mathbb{R}^{B\times 1\times 2}\\
\boldsymbol{S} &= \boldsymbol{a}_{\mathcal{I}} \cdot \boldsymbol{t}_{\mathcal{I}} + \boldsymbol{a}_{\mathcal{L}} \cdot  \boldsymbol{t}_{\mathcal{L}},\quad \boldsymbol{S} \in \mathbb{R}^{B\times d}
\end{aligned}
\end{equation}

{\flushleft \textbf{Vision-Language Modality Fusion.}}
The goal of fusion is to combine multiple modalities to leverage the complementarity of heterogeneous image and language data and provide more robust predictions. Therefore, we employ a modality-wise fusion to make multimodal interdependencies to achieve better generalization of FAS. Fig.~\ref{CGM}(2) shows the vision-language modality fusion. Formally, given textual feature vectors $\boldsymbol{S} \in \mathbb{R}^{B\times d}$ and visual feature vectors $\boldsymbol{V} \in \mathbb{R}^{B\times d}$ in different modalities, we first utilize two projection matrix $W_t, W_v\in \mathbb{R}^{d \times n}$ (denoted as $\textit{fc}(\cdot)$ in Fig.~\ref{CGM}) to map the two input features to a lower dimensional space, respectively. Then we transform these two dimensionality-reduced feature vectors to corresponding circulant matrices, devoted as $\mathcal{R}(\cdot)$. Study~\cite{fukui2016multimodal,yu2017multi} has emphasized that comprehensively harnessing the interactions among elements of multi-modal features can significantly bolster performance outcomes. In order to make elements in the projection vector and the circulant matrix fully interact, we explore matrix multiplication between the circulant matrix and the projection vector. And we avoid introducing new parameters or increasing the computational cost of multi-modal fusion. This process can be expressed as follows:

\begin{equation}
\begin{aligned} 
\boldsymbol{S}^{\prime} &=\mathcal{R}(S W_t),\quad \boldsymbol{S}^{\prime} \in \mathbb{R}^{B \times n\times n}\\
\boldsymbol{V}^{\prime} &= \mathcal{R}(\boldsymbol{V} W_v),\quad \boldsymbol{V}^{\prime} \in \mathbb{R}^{B \times n\times n}\\
\boldsymbol{Z} &=\boldsymbol{S}^{\prime} S W_t + \boldsymbol{V}^{\prime} \boldsymbol{V} W_v,\quad \boldsymbol{Z} \in \mathbb{R}^{B \times n} \\
\end{aligned}
\end{equation}

Subsequently, the final fused features $\boldsymbol{Z}$ are passed through a projector $\mathcal{F}(\cdot) $ to obtain the final predictions. The classification loss is computed as follows:

\begin{equation}
\mathcal{L}_{c l s}=\operatorname{CE}(\mathcal{F}(Z), y))
\end{equation}
where $\operatorname{CE}(\cdot)$ donates the Cross-Entropy loss $\mathcal{F}(\cdot)$ consists of two linear layers followed by a softmax function, and $y$ is the corresponding label to the image.

In the training stage, the two text encoders are frozen, and others, including Q-former and CGM, are learnable. The overall loss can be formulated as follows:
\begin{equation}
\mathcal{L}_{}=\mathcal{L}_{\mathrm{cls}}+ \mathcal{L}_{\mathrm{kg}}
\end{equation}

\section{Experiments}

\begin{table*}[h]
  \centering
  \caption{Protocol 1 results on Idiap-Replay-Attack (\textbf{I}), CASIA-FASD (\textbf{C}), MSU-MFSD (\textbf{M}) and OULU-NPU (\textbf{O}) datasets.  Note that the $*$ indicates the corresponding method using CelebA-Spoof as the supplementary source dataset. The \textbf{bold} numbers highlight the best performance.}
  \begin{tabular}{ >{\arraybackslash}p{3.9cm}
  >{\centering\arraybackslash}p{1.05cm}
  >{\centering\arraybackslash}p{1.05cm}
  >{\centering\arraybackslash}p{1.05cm}
  >{\centering\arraybackslash}p{1.05cm}
  >{\centering\arraybackslash}p{1.05cm}
  >{\centering\arraybackslash}p{1.05cm}
  >{\centering\arraybackslash}p{1.05cm}
  >{\centering\arraybackslash}p{1.05cm}
  >{\centering\arraybackslash}p{1.05cm}}
  \toprule
  \multicolumn{1}{l}{Method} &\multicolumn{2}{c}{OCI$\rightarrow$M}  & \multicolumn{2}{c}{OMI$\rightarrow$C} & \multicolumn{2}{c}{OCM$\rightarrow$I} & \multicolumn{2}{c}{ICM$\rightarrow$O} & {avg.} \\
    \cmidrule{2-10}        & HTER(\%)      & AUC(\%)        & HTER(\%)        & AUC(\%)        & HTER(\%)        & AUC(\%)        & HTER(\%)        & AUC(\%)       & HTER(\%)         \\
  \midrule
MP(TIFS'22)~\cite{cai2022learning} & 5.24    & 97.28      & 9.11     &   96.09    &  15.35     & 90.67    & 12.40    &   94.26   & 10.53 \\
CIFAS(ICME'22)~\cite{liu2022causal}   &5.95    &96.32       &  10.66     &  95.30    &  8.50     & 97.24    & 13.17   &   93.44    &     9.56\\

PatchNet(CVPR'22)~\cite{wang2022patchnet}           & 7.10        & 98.46      & 11.33       & 94.58      & 13.40       & 95.67      & 11.82       & 95.07      & 10.91           \\
NDA-C(TIFS'23)~\cite{wang2023domain} & 4.29     &99.27       &  13.22    &  92.89    &  5.50     &   96.90    & 17.45    &   89.93    &  10.12  \\
CRFAS(ICASSP'23)~\cite{zheng2023learning} & 6.90    &98.30      &  9.33     &   97.31    &  7.78      &   97.14    & 15.52    &   92.18   &  9.88  \\
UDG-FAS(ICCV'23)~\cite{liu2023towards}     &  7.14    &   97.31    &    11.44    & 95.59      &  6.28      &  98.61     &    12.18    &  94.36     &   9.62    \\
SA-FAS(CVPR'23)~\cite{sun2023rethinking}          & 5.95        & 96.55      & 8.78        & 95.37      & 6.58        & 97.54      & 10.00       & 96.23      & 7.82              \\
IADG(CVPR'23)~\cite{zhou2023instance}               & 5.41        & 98.19      & 8.70        & 96.44      & 10.62       & 94.50      & 8.86        & 97.14      & 8.39       \\
DiVT-M(WACV'23)~\cite{liao2023domain}      &  2.86     & 99.14    &   8.67     &  96.92    &  3.71     &  99.29   &  13.06    & 94.04     & 7.08        \\
ViT-C\&FA\&CS(IJCV'24)~\cite{cai2024towards} & 4.62   & 98.92 & 7.28  &97.02   & 10.89  & 97.05  &6.77 &98.25  &7.39  \\
S-Adapter(TIFS'24)~\cite{cai2024s}      &  2.90     & 99.48    &   7.37     &  97.63    &  8.54     &  97.17   &  8.20    & 97.69    & 6.75     \\
CA-MoEiT(IJCV'24)~\cite{liu2024moeit}   & 2.88     & 98.76     & 7.89   & 97.70  & 6.18    & 98.94  & 9.72  & 96.22   &6.67 \\
\midrule
CLIP-V(PMLR'21)~\cite{radford2021learning}                 & 4.29  & 98.76 & 5.00  & 98.89 & 7.14  & 97.92 & 6.09  & 98.12 & 5.63   \\
CLIP(PMLR'21)~\cite{radford2021learning}               & 4.04  & 99.13 & 5.00  & 98.89 & 6.57 & 98.45 & 6.09  & 98.12 & 5.43   \\
CoOp(IJCV'22)~\cite{zhou2022coop}                & 3.86  & 99.08 & 2.33  & 98.92 & \textbf{6.07}  & 98.52 & 5.83  & 98.97 & 4.37 \\
CoCoOp(CVPR'22)~\cite{zhou2022cocoop}      &            4.16& 99.01  &5.17    &98.19  &6.21  &98.50  & 6.00  & 98.49 & 5.39 \\
\cellcolor{gray!20}\textbf{CCPE}(Ours)     & \cellcolor{gray!20}\textbf{3.10}    & \cellcolor{gray!20}\textbf{99.21}    &  \cellcolor{gray!20}\textbf{1.33}        & \cellcolor{gray!20}\textbf{99.36}     & \cellcolor{gray!20}6.08      &  \cellcolor{gray!20}\textbf{94.36}      &  \cellcolor{gray!20}\textbf{5.57}        &  \cellcolor{gray!20}\textbf{98.49}      &  \cellcolor{gray!20}\textbf{4.02}   \\
\midrule
ViTAF$^*$-5-shot(ECCV'22)~\cite{huang2022adaptive}      & 2.92   & \textbf{99.62 } &1.40  &99.92 &\textbf{1.64} &\textbf{99.64}  &5.39   &98.67  &3.31  \\
FLIP-MCL$^*$(ICCV'23)~\cite{srivatsan2023flip}      & 4.95    &98.11   & 0.54   &\textbf{99.98} &4.25  &99.07  &\textbf{2.31}   &99.63  &3.01  \\
\cellcolor{gray!20}\textbf{CCPE$^*$}(Ours)  & \cellcolor{gray!20}\textbf{2.86}    & \cellcolor{gray!20}99.24    &  \cellcolor{gray!20}\textbf{1.30}        & \cellcolor{gray!20}\textbf{99.98}     &
\cellcolor{gray!20}   4.15  & 
\cellcolor{gray!20}99.31      &  \cellcolor{gray!20}3.64        &  \cellcolor{gray!20}\textbf{99.67}      &  \cellcolor{gray!20}\textbf{2.99}   \\
\bottomrule
\end{tabular}
\label{tab:icmo_results}
\end{table*}

\subsection{Experiments Settings}

{\flushleft \textbf{Experiment Datasets \& Protocols.}}
Following the prior works~\cite{srivatsan2023flip}, two varied protocols are employed to evaluate the model’s generalization performance. In Protocol 1, we utilize four benchmark datasets, including \textbf{I}diap-Replay-Attack(\textbf{I})~\cite{chingovska2012effectiveness}, \textbf{C}ASIA-FASD(\textbf{C})~\cite{zhang2012face}, \textbf{M}SU-MFSD (\textbf{M})~\cite{wen2015face} and \textbf{O}ULU-NPU(\textbf{O})~\cite{Boulkenafet2017OULU} on cross-dataset testing. In \textit{Protocol 2}, we evaluate the effectiveness on the large-scale FAS datasets, CASIA-\textbf{S}URF(\textbf{S})~\cite{zhang2020casia}, CASIA-SURF \textbf{C}eFA(\textbf{C})~\cite{liu2021casia} and \textbf{W}MCA(\textbf{W})~\cite{george2019biometric}. In addition, we also report the experimental results of using CelebA-Spoof~\cite{CelebA-Spoof} as a supplementary source dataset. In each protocol, we regard each dataset as one domain and use the Half Total Error Rate (HTER) and Area Under Curve (AUC) as evaluation metrics~\cite{zhou2022generative}. The performance metrics obtained from the follow-up evaluation represent a relatively stable result, which has been consistently observed after conducting a series of rigorous experiments. 

{\flushleft \textbf{Implementation Details.}}
We pre-process all face images to $224 \times 224 \times 3$ and split them into a patch size of $14 \times 14$. The image and text encoders are adapted from pre-trained ViT-B/16 of CLIP. For the learnable content prompts, the number of queries is 8. The depth of the Q-former is set to 1. Each feature vector extracted by encoders has a dimension of 512. In our approach, PyTorch serves as the foundation for implementation, complemented by training via the Adam optimizer. This process involves an initial learning rate set to $10^{-6}$ and a batch size maintained at 32.

\subsection{Comparisons to Prior SOTA Results}
To illustrate our model's ability to adapt to unseen domains, we employ Leave-One-Out (LOO) validation and give the result summarized in Tab.~\ref{tab:icmo_results}. This process involves executing cross-domain in four typical scenarios for the FAS task. In this framework, we randomly choose three datasets as source domains, and the fourth dataset is designated as the unseen target domain. Tab.~\ref{tab:icmo_results} states the comparison of these methods in two parts: conventional approaches for direct manipulation of image features and CLIP-based prompt engineering techniques. Notably, all the results in Tab.~\ref{tab:icmo_results} are derived without employing the CelebA-Spoof~\cite{CelebA-Spoof} dataset as an additional resource.

From Tab.~\ref{tab:icmo_results}, we have the following observations. (1) Under these four cross-dataset benchmarks, conventional DG FAS approaches demonstrate subpar performance. Specifically, on the HTER, CCEP improves the best baseline S-Adapter by 6.07\% in the OMI$\rightarrow$C, 2.63\% in the ICM$\rightarrow$O, 2.46\% in the OCM$\rightarrow$I and improves the subpar baseline DiVT-M~\cite{liao2023domain} by 7.34 \% in the setting OMI$\rightarrow$C, 7.49 \% in the setting ICM$\rightarrow$O. Compared to the baseline IADG~\cite{zhou2023instance}, which allows generalization of features by weakening the sensitivity of features to instance-specific styles, CCPE shows outstanding improvements of 1.31\% in the OCI$\rightarrow$M setting, 7.47\% in OMI$\rightarrow$C, 4.54\% in OCM$\rightarrow$I, and 0.52\% in ICM$\rightarrow$O. The reason is that these models are guided only by the image data and the corresponding image labels during training. Such an operation imposes the limitation of insufficient representative ability. (2) Our method outperforms most CLIP-based methods. Compared to CLIP and CoOp, our method has improved by 1.41\%, 0.25\% in terms of average HTER. This validates our idea of using content features to assist prompt learning can learning generalized representations in FAS. (3) By incorporating the CelebA-Spoof~\cite{CelebA-Spoof} dataset into training data, our CCPE$^*$ maintains the best performance compared to competitors ViTAF-5-shot$^*$~\cite{huang2022adaptive} and FLIP-MCL$^*$~\cite{srivatsan2023flip}, achieving the best average HTER of 2.99\%.

In Tab.~\ref{tab:scw_results}, we compare the results of CCPE with several baseline methods, such as ViT\cite{huang2022adaptive}, CLIP-V~\cite{radford2021learning}, CLIP~\cite{radford2021learning}, CoOp~\cite{zhou2022coop}, and CoCoOp~\cite{zhou2022cocoop} on Protocol 2, where CLIP-V indicates removing the text encoder from CLIP. Our CCPE significantly outperforms the baseline methods in all sub-protocols, with the HTER results decreasing from 12.00\%, 15.12\%, and 10.46\% to 9.57\%, 14.89\%, and 8.16\%, respectively, for the target domains CS$\rightarrow$W, SW$\rightarrow$C, and CW$\rightarrow$S, compared to CoOp. The average HTER of CCPE is 10.87\%, which is lower than all the baseline methods. Interestingly, the performance of CoCoOp is worse than CoOp, with average HTER of 14.38\% and 12.53\%, respectively. While CoCoOp generally performs better than CoOp in recognition tasks with specific semantic categories, we analyze that in FAS tasks, the ambiguous semantic categories, such as the category descriptions, have no semantic information for CLIP, leading to CoCoOp's suboptimal performance than CoOp. When introducing the CelebA-Spoof~\cite{CelebA-Spoof} dataset into the training data, our CCPE$^*$ maintains the best performance compared to competitors, achieving an HTER of 4.42\% and an AUC of 99.30\% for the target domain W, outperforming ViT$^*$\cite{huang2022adaptive} and FLIP-MCL$^*$\cite{srivatsan2023flip}. CCPE$^*$ demonstrates superior performance compared to ViT$^*$ and FLIP-MCL$^*$, achieving a lower average HTER of 8.33\%. In contrast, ViT$^*$ and FLIP-MCL$^*$ yield higher HTER of 10.82\% and 8.61\%, respectively, highlighting the effectiveness of the CCPE$^*$ approach in reducing misclassifications when facing DG problem.

\begin{table*}[htbp]
  \centering
\caption{Protocol 2 results on CASIA-SURF (S), CASIA-SURF CeFA (C), and WMCA (W) datasets. Note that the $*$ indicates the corresponding method using CelebA-Spoof as the supplementary source dataset. Bold numbers highlight the best performance.}
  \begin{tabular}{ >{\arraybackslash}p{2.7cm}
   >{\centering\arraybackslash}p{1.4cm}
  >{\centering\arraybackslash}p{1.4cm}
  >{\centering\arraybackslash}p{1.4cm}
  >{\centering\arraybackslash}p{1.4cm}
  >{\centering\arraybackslash}p{1.4cm}
  >{\centering\arraybackslash}p{1.4cm}
  >{\centering\arraybackslash}p{1.4cm}}
  \toprule
  \multicolumn{1}{c}{Method} &\multicolumn{2}{c}{CS$\rightarrow$W}  & \multicolumn{2}{c}{SW$\rightarrow$C} & \multicolumn{2}{c}{CW$\rightarrow$S} & {avg.} \\
    \cmidrule{2-8}         & HTER(\%)      & AUC(\%)        & HTER(\%)        & AUC(\%)            & HTER(\%)        & AUC(\%)       & HTER(\%)         \\
  \midrule
ViT~\cite{huang2022adaptive}              & 22.18    & 89.76   & 17.59  & 89.71 & 17.11  & 90.46    & 18.96  \\
CLIP-V~\cite{radford2021learning}             & 21.88    & 88.49   & 17.00  & 90.24 & 17.05  & 92.97    &  18.64  \\
CA-MoEiT~\cite{liu2024moeit}  & 18.49                 & 90.08                & 19.25                 & 89.93                & 15.56                 & 91.01                & 17.77                 \\
CLIP~\cite{radford2021learning}               & 16.74    & 89.99   & 15.31   & 88.75 & 14.01  & 96.45    &   15.35 \\
CoOp~\cite{zhou2022coop}                 & 12.00    & 93.74   & 15.12  &89.05  &10.46  & 96.73  &   12.53  \\
CoCoOp~\cite{zhou2022cocoop}          & 13.89    & 90.74 &  15.49 & 89.40 & 13.76 & 95.59 & 14.38  \\
\cellcolor{gray!20}\textbf{CCPE(Ours)}       & \cellcolor{gray!20}\textbf{9.57}    & \cellcolor{gray!20}\textbf{94.25}    & \cellcolor{gray!20}\textbf{14.89}        &\cellcolor{gray!20}\textbf{91.56}     &\cellcolor{gray!20}\textbf{8.16}      & \cellcolor{gray!20}\textbf{96.78}           &\cellcolor{gray!20} \textbf{10.87}             \\
\midrule
 $\text{ViT}^*$~\cite{huang2022adaptive}                             & 7.98  & 97.97 & 11.13  & 95.46 & 13.35  & 94.13 & 10.82   \\
 $\text{FLIP-MCL}^*$~\cite{srivatsan2023flip}             &4.46   &99.16 & 9.66 &96.69  & 11.71  & 95.21 &  8.61 \\

\cellcolor{gray!20} \textbf{$\text{CCPE}^*$(Ours)}      &\cellcolor{gray!20} \textbf{4.42}    &\cellcolor{gray!20} \textbf{99.30}    &\cellcolor{gray!20} \textbf{9.59}        &\cellcolor{gray!20}\textbf{96.73}     &\cellcolor{gray!20}\textbf{10.97}      &\cellcolor{gray!20} \textbf{97.40}           & \cellcolor{gray!20}\textbf{8.33}             \\

\bottomrule
\end{tabular}
\label{tab:scw_results}
\end{table*}

\begin{table}[htbp]
  \centering
\caption{Ablation study of each component. The components were evaluated the Inherent Content Prompt Generation (\textit{ICPG}), Learnable Content Prompt Generation (\textit{LCPG}) and the Cross-modal Guidance Module (\textit{CGM}).}
  \begin{tabular}{
  >{\centering\arraybackslash}p{0.55cm}
  >{\centering\arraybackslash}p{0.55cm}
  >{\centering\arraybackslash}p{0.55cm}|
  >{\centering\arraybackslash}p{0.9cm}
   >{\centering\arraybackslash}p{0.9cm}
    >{\centering\arraybackslash}p{0.9cm}
     >{\centering\arraybackslash}p{0.9cm}}
  \toprule
     \multicolumn{3}{c}{Components}&  \multicolumn{2}{c}{OMI$\rightarrow$C}  &  \multicolumn{2}{c}{ICM$\rightarrow$O} \\
 \midrule
 \textit{ICPG}& \textit{LCPG} & \textit{CGM}      & HTER(\%)   & AUC(\%) & HTER(\%)   & AUC(\%)      \\
  \midrule
 \checkmark & &                 & 4.62  &  97.57   &7.01 & 97.93    \\
          & \checkmark&           & 3.35  &  97.94 &5.94 & 98.21    \\
 \checkmark& \checkmark&                  & 1.64  & 99.18 &5.69 &98.33 \\   

\cellcolor{gray!20}\checkmark&\cellcolor{gray!20} \checkmark &  \cellcolor{gray!20}\checkmark     &\cellcolor{gray!20}\textbf{1.33} & \cellcolor{gray!20}\textbf{99.36} &\cellcolor{gray!20}\textbf{5.57} & \cellcolor{gray!20}\textbf{98.49}  \\
    \bottomrule
    \end{tabular}

\label{tab:component_effect}
\end{table}

\begin{figure*}[ht]
\centering
\includegraphics[width=0.98\linewidth]{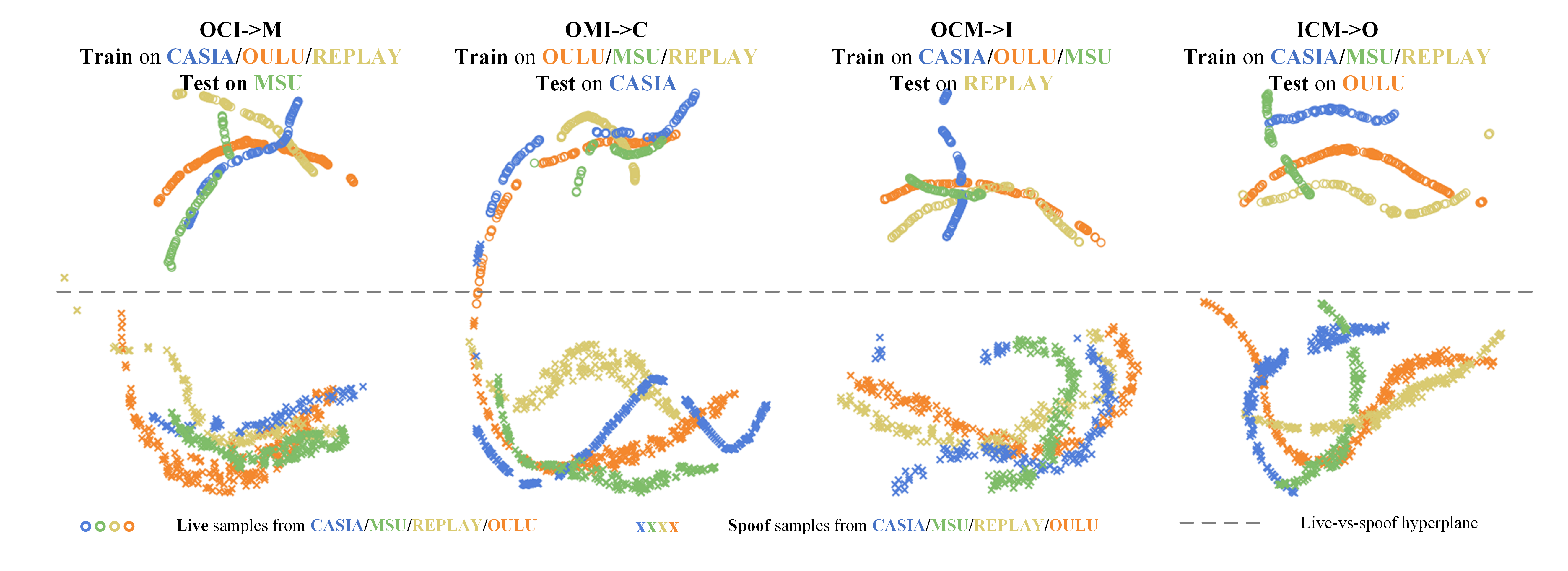}
\caption{UMAP~\cite{umap2018} 
 visualization for the feature learned from the penultimate layer of the proposed CCPE method in the cross-dataset FAS task of \textit{Protocol 1}. The dotted line in each visualization represents the decision boundary derived from the training samples in the 2D space. The consistent separation of live and spoof samples across different domain combinations highlights the ability of CCPE to learn domain-invariant features for FAS.}
\label{umap}
\end{figure*}

\subsection{Ablation Studies}
{\flushleft \textbf{Effectiveness of each component.}} We conducted an ablation study to assess the individual contributions of each component in our proposed framework: the Inherent Content Prompt Generation (\textit{ICPG}), Learnable Content Prompt Generation (\textit{LCPG}) and the Cross-modal Guidance Module (\textit{CGM}).
The experimental results of various combinations on OMI$\rightarrow$C and ICM$\rightarrow$O are listed in Tab.~\ref{tab:component_effect}. In the first row, we just employ the instruction-based LLM to automatically semantically generate rich text descriptions semantic description. Regrettably, they exhibit subpar performance when dealing with unseen domains. The performance of learnable prompt with content-aware queries is slightly better than the fixed text descriptions, i.e., for in HTER, improve 1.27\% in OMI$\rightarrow$C and 1.07\% in ICM$\rightarrow$O. Then we set a composite prompt that includes inherent prompt benefits from both explicitly and implicitly extracting the most informative visual content. This composite prompt structure plays an indispensable role in the application of prompt engineering because fixed template prompts are good at describing inherent attack types, while learnable context prompts can better cover unknown attack types. Constructing composite prompts combines the advantages of both, and can better solve the difficulty of diverse attacks in FAS tasks. 
Moreover, \textit{CGM} enhances deep interactivity on the language and vision branches. \textit{CGM} achieves 0.31\% and 0.12\% improvement of OMI$\rightarrow$C and ICM$\rightarrow$O settings. In summary, the goal of all components is to assist the base network in learning the complete and robust spoof feature. So the last row of our whole CCPE achieves the best results in various cases.

{\flushleft \textbf{Effectiveness of CGM.}} 
Tab.~\ref{tab:fusion_effect} compares the proposed Cross-modal Guidance Module (CGM) with baseline fusion methods, including Element-wise sum, Element-wise Product, Concatenation, and Transformer~\cite{dosovitskiy2020image} in the cross-dataset FAS task. CGM outperforms all baselines in both OMI$\rightarrow$C and ICM$\rightarrow$O settings, achieving the lowest HTER and highest AUC. These results demonstrate CGM's effectiveness in capturing complementary information from different modalities and generating discriminative features that generalize well to unseen domains, highlighting the importance of guided attention and cross-modal interactions in our method. Even though we use a 1-block transformer structure in order to reduce the number of parameters, the model struggles with fine-tuning due to the excessive number of parameters in its structure, resulting in performance lower than CGM and even concatenation.

\begin{table}[htbp]
  \centering
\caption{Ablation study of fusion module. The fusion modules evaluated were the Element-wise sum, Element-wise product, Concatenation, Transformer, and Cross-modal Guidance Module (\textit{CGM}).}
 \begin{tabular}{>{\centering\arraybackslash}p{3 cm}|
  >{\centering\arraybackslash}p{0.9cm}
  >{\centering\arraybackslash}p{0.9cm}
  >{\centering\arraybackslash}p{0.9cm}
   >{\centering\arraybackslash}p{0.9cm}}
\toprule
\multicolumn{1}{c}{}& \multicolumn{2}{c}{OMI$\rightarrow$C} & \multicolumn{2}{c}{ICM$\rightarrow$O} \\
\midrule
Fusion Module & HTER(\%) & AUC(\%) & HTER(\%) & AUC(\%) \\
\midrule
\textit{Element-wise Sum} & 4.27 & 96.08 & 7.64 & 97.92 \\
\textit{Element-wise Product} & 3.95 & 96.35 & 7.31 & 98.06\\
\textit{Concatenation}  & 1.64  & 99.18 &6.14 &98.01 \\
\textit{Transformer} & 2.05 & 97.94 &6.48 & 98.35\\

\cellcolor{gray!20}\textit{CGM} &\cellcolor{gray!20}\textbf{1.33} & \cellcolor{gray!20}\textbf{99.36} &\cellcolor{gray!20}\textbf{5.57} & \cellcolor{gray!20}\textbf{98.49} \\
\bottomrule
\end{tabular}
\label{tab:fusion_effect}
\end{table}

\begin{table}[h]
  \centering
  \caption{Ablation study of different LLM modal. The LLM models were evaluated the Opt2.7b, MiniGPT-4, and LLaVA.}
  \begin{tabular}{@{}c *{2}{>{\centering\arraybackslash}p{1.3cm}} *{2}{>{\centering\arraybackslash}p{1.3cm}}@{}}
    \toprule
    \multirow{2}{*}{LLM} 
    & \multicolumn{2}{c}{Simple Instruction} 
    & \multicolumn{2}{c}{Complicated Instruction} \\
    \cmidrule(lr){2-3} \cmidrule(lr){4-5} 
    & HTER (\%) & AUC (\%) & HTER (\%) & AUC (\%) \\ 
    \midrule
    Opt2.7b~\cite{zhang2022opt} & 6.02 & 97.97 & 7.32 & 97.86 \\
    MiniGPT-4~\cite{zhu2023minigpt} & 5.69 & 98.51 & 5.74 & 98.83 \\
    LLaVA~\cite{liu2024visual} & 5.57 & 98.49 & 5.70 & 98.79 \\ 
    \bottomrule
  \end{tabular}
  \label{tab:different LLM}
\end{table}

\begin{table}[htbp]
\centering
\caption{Ablation of the Number of queries and the depth for Q-former. The optimal value is elegantly highlighted in bold. Excessive depth and numbers may cause overfitting or complicate optimization. Thus, a Q-former depth of 1 and query number of 8 seems to be the optimal balance between efficiency and effectiveness.}
\begin{tabular}{
  >{\centering\arraybackslash}p{2.5cm}|
  >{\centering\arraybackslash}p{0.9cm}
  >{\centering\arraybackslash}p{0.9cm}
  >{\centering\arraybackslash}p{0.9cm}
  >{\centering\arraybackslash}p{0.9cm}}
\toprule
   HTER(\%)   & \multicolumn{4}{c}{Number of queries} \\
\midrule
Q-former Depth &    8      &   16      &   32      &   64      \\
\midrule
$\times \: 1$      & \textbf{5.57}    &    7.08      &    8.65    &  6.44       \\
$\times  \: 2$      & 6.31             &   5.86       &    8.32    &  6.50      \\
$\times  \: 4$      & 6.29             &   6.51       &    6.74    &  8.03      \\
$\times  \: 8$      & 5.98             &   7.13       &    7.53    &  7.25        \\
 \bottomrule
\end{tabular}

\label{tab:query_depth}
\end{table}

{\flushleft \textbf{Different LLM modal.}} Our method can flexibly utilize various LLMs and is not limited to the models tested in our current experiments. This adaptability ensures robustness and future applicability as newer LLMs become available. Tab.~\ref{tab:different LLM} presents an ablation study comparing three LLMs (LLaVA~\cite{liu2024visual}, MiniGPT-4~\cite{zhu2023minigpt}, and Opt2.7b~\cite{zhang2022opt}) on the task of ICM$\rightarrow$O. We evaluated two instruction types: a simple instruction ("Summarize the visual content of the image in one sentence.") and a complicated instruction ("Please describe these images based on several attributes: local consistency and clarity of pixels, contours of the face and facial features, harmony of facial features, gloss and reflection effects, sense of depth and three-dimensionality, and the presence of phone screens or paper edges."). The results indicate that MiniGPT-4 and LLaVA consistently outperform Opt2.7b, achieving lower HTER and higher AUC under both instruction types. Comparing MiniGPT-4 and LLaVA, LLaVA shows slightly lower HTER scores for both instructions, suggesting better overall performance, whereas MiniGPT-4 achieves slightly higher AUC under complicated instructions. Additionally, complicated instructions generally yield slightly worse performance across all evaluated models, possibly due to increased complexity, subjectivity, and ambiguity in detailed attribute descriptions.

{\flushleft \textbf{Different number of queries and the depth of Q-former.}} 
We assessed how the number of queries and the depth of Q-former affects performance, as detailed in Tab.~\ref{tab:query_depth}. The search for the optimal value is carried out within the experiment $\mathrm{OMI} \rightarrow \mathrm{C}$. Our findings indicate that the Q-former depth is 1 while queries with a number of 8 deliver the most consistent results. However, a greater number do not improve the performance of the learnable query prompt. A depth of 8 consistently falls short in all scenarios, suggesting that excessive depth may cause overfitting or complicate optimization. Thus, a Q-former depth of 1 and the query number of 8 seem to be the optimal balance between efficiency and effectiveness.

\subsection{Visualization and Analysis}
{\flushleft \textbf{UMAP visualization.}} 
Fig.~\ref{umap} presents UMAP\cite{umap2018} visualizations of the feature space learned from the penultimate layer of CCPE in the cross-dataset FAS task of \textit{Protocol 1}. The figure employs a color scheme to distinguish between the four datasets: CASIA-FASD (CASIA) in blue, MSU-MFSD (MSU) in green, Idiap-Replay-Attack (REPLAY) in yellow, and OULU-NPU (OULU) in orange. This color-coding remains consistent across all model plots, allowing for easy identification of the dataset to which each data point belongs. A distinct hyperplane separating the live and spoof clusters is consistently observed across all domain combinations, suggesting that CCPE learns domain-invariant features that can generalize effectively to unseen domains. In the OCM$\rightarrow$I setting, the CASIA, OULU-NPU(OULU), and MSU-MFSD(MSU) training samples form distinct clusters, indicating the presence of domain-specific characteristics. However, despite these domain shifts, the live-vs-spoof hyperplane learned by CCPE remains consistent and effectively separates the live and spoof samples from the unseen REPLAY test set. This observation, along with the consistent separation of live and spoof samples in other cross-dataset settings (OCI$\rightarrow$M, OMI$\rightarrow$C, and ICM$\rightarrow$O), demonstrates that CCPE captures the intrinsic differences between live and spoof samples, rather than overfitting to domain-specific patterns.

\begin{figure}[t]
\centering
\includegraphics[width=1\linewidth]{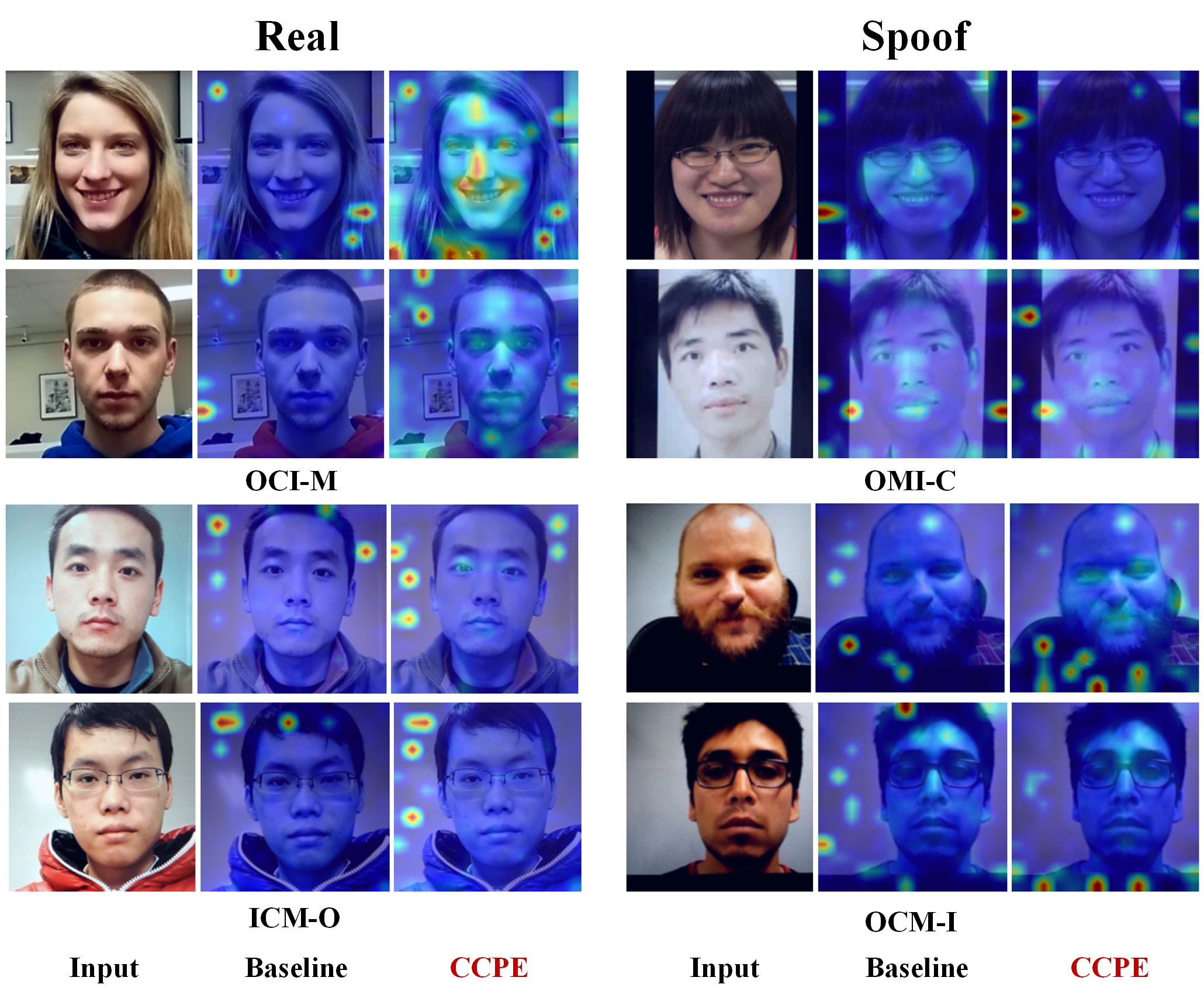}
\caption{Visualization of attention maps on images from different scenarios in \textit{Protocol 1}. Classification errors arose in the baseline model due to its shortcomings in detecting spoofing regions. Our CCPE accurately adjusts the region of interest by excluding domain-related interfering information and excels in correctly classifying these samples.}

\label{Visualization}
\end{figure}

{\flushleft \textbf{Attention map visualization.}} 
 Utilizing the attention-model explainability tool~\cite{chefer2021generic}, a comparative visual analysis was performed to demonstrate the superior performance of the proposed CCPE compared to the baseline. Fig.~\ref{Visualization} systematically presents these findings, where it is evident that the visual attention maps of the baseline are associated with classification inaccuracies, in stark contrast to the precise classification capabilities exhibited by the CCPE approach. Concretely, the left side of Fig.~\ref{Visualization} shows the real face images, while the right side shows the spoof images. The first column on each side refers to the input images and the second column is the corresponding activation maps through the baseline model. The baseline contains only the built-in structure of the clip, which is equipped with manual category-based prompts. 
The proposed CCPE emphasizes the internal regions of genuine human faces as classification cues, while for attack images, the focus is more scattered across attack indicators. The baseline model's classification errors arise from its shortcomings in detecting spoofing regions, while our CCPE accurately adjusts the region of interest and excels in correctly classifying these samples.

\section{Conclusion}
In this work, we propose CCPE, an efficient prompt engineering framework for adapting VLMs to generalizable FAS. We underline the importance of prompt engineering to achieve complex tasks (i.e. DG FAS). Through the content-aware composite prompt generation module, we form both explicit and implicit content prompts that carry instance-specific visual content. In addition, we design the cross-model guidance module to adaptively fuse uni-modal branches and align the visual-language representation to produce state-of-the-art performance on FAS classification.

\ifCLASSOPTIONcaptionsoff
  \newpage
\fi
\bibliographystyle{IEEEtran}
\bibliography{IEEEabrv,CCPE}

\end{document}